# Machine Learning application in Health

Ghadah Alshabana
galshaba@gmu.edu
George Mason University

Marjn Sadati
msdati@gmu.edu
George Mason University

Thao Tran
ttran81@gmu.edu
George Mason University

Michael Thompson George
mthomp19@gmu.edu
George Mason University

Ashritha Chitimalla
achitima@gmu.edu
George Mason University

*Abstract*—Coronavirus can be transmitted through the air by close proximity to infected persons. Commercial aircraft are a likely way to both transmit the virus among passengers and move the virus between locations. The importance of learning about where and how coronavirus has entered the United States will help further our understanding of the disease. Air travelers can come from countries or areas with a high rate of infection and may very well be at risk of being exposed to the virus. Therefore, as they reach the United States, the virus could easily spread. On our analysis, we utilized machine learning to determine if the number of flights into the Washington DC Metro Area had an effect on the number of cases and deaths reported in the city and surrounding area.

## I. Introduction

The COVID-19 pandemic has caused a serious impact on the world economy, with many deaths and long-term injuries across the world, an increase in businesses bankruptcy with an associated increase in lost jobs, and an increase in food scarcity as well. Moreover, health care systems, as well as the airline industry, faced enormous financial challenges as 30% of the United States airlines' stock prices decreased during this crisis [1]. During this crisis, many airlines canceled their flights and applied travel restrictions to control the spread of coronavirus, leading to a significant impact on the airline sector. According to [2], [3], the ongoing COVID-19 pandemic is the gravest crisis that happened to the aviation sector, and it may take around five years to recover from it. As thousands of employees were released, international tourism was frequently stopped, and global quarantine restrictions limited the flights that continued. [4]–[7] [8], [9] [10]–[13] [14]–[16] [17]–[22] [23]–[25]

Given the serious effect COVID-19 has had on the world, there is a significant necessity to understanding how the novel coronavirus is transmitted and the various factors that allow it to rapidly disperse through a community, city, or nation. Among the risk factors to a given location, air travel may well be a factor, or have historically played a factor in allowing additional spread of the coronavirus from more infected regions to those with limited or no prior infections.

Numerous airplanes are supported with High Efficiency Particulate Air (HEPA) filters; however, the filter's effectiveness only applies to the air that goes through it, and it does not necessarily filter every onboard virus; therefore, airlines require people to wear masks during the flight [26]. Additionally, close contact between people is the prime cause of spreading the COVID-19 easily [27] which led many airlines to leave the middle seat empty in their aircraft [28]. However, with all travel restrictions and procedures, COVID-19 can still be transmitted between people on airplanes, as stated by [29]–[31]. Therefore, our aim is to discover the correlation, if present, between the number of flights and resulting coronavirus cases in the DC area.

The importance of learning about where and how corona virus has entered the United States will help further our understanding of the disease. According to CDC [32], the first coronavirus case in the US has been identified in Washington state, and that was due to air travel from Wuhan, China. The most common way COVID-19 can spread is by human interaction, through respiratory droplets such as talking, coughing, sneezing, and more. Air travelers can come from countries or areas with a high rate of infection and may very well be at risk of being exposed to the virus. Therefore, as they reach the United States, the virus could easily spread. In our analysis, we intend to use the OpenSky dataset records and combine it with CDC data to determine if the number of flights into or out of the Washington DC metro area may have impacted the number of coronavirus deaths reported in those counties and the region surrounding the respective airports in question.

Other analyses have concluded that coronavirus can travel via flight and there is an inverse relationship between distance to an airport and how many coronavirus cases result from travel into the region. We suspect that the District of Columbia will show different results as a significant portion of business and political activity in the region is focused on the US federal government.

## II. Datasets

In this paper, we have utilized two dataset sources. The flight dataset was obtained from OpenSky, showing the air traffic during the coronavirus pandemic. Table I summarizes the OpenSky attributes [33]. The coronavirus dataset was obtained from the New York Times and shows the number of cases and death in the United States. Table II summarizes the New York Times [34].

TABLE I
OPENSKY ATTRIBUTES

| Variable Name | Description | Type |
|---|---|---|
| Callsign | the identifier of the flight | String |
| Registration | the aircraft tail number | String |
| Origin | the origin flight airport represented with four letters. | String |
| Destination | the destination flight airport, represented with four letters. | String |
| Firstseen | UTC timestamp of the first message received by the OpenSky Network. | String |
| Lastseen | UTC timestamp of the last message received by the OpenSky Network. | String |

TABLE II
NEW YORK TIMES ATTRIBUTES

| Variable Name | Description | Type |
|---|---|---|
| Date | the date of the reported Covid-19 cases and deaths. | Date |
| State | the name of the state. | String |
| County | the name of the county. | String |
| Fips | standard geographic identifier. | Number |
| Cases | the total number of cases of Covid-19. | Number |
| Deaths | the total number of deaths from Covid-19. | Number |

As for the cleaning process. From the OpenSky dataset: the destination variable (filtering for Baltimore International Airport, Dulles International Airport, and Reagan National Airport), and Last seen variable as that gave us an indication of the date the flight was occurring. From the New York times dataset, we used the date, state, and county to filter down to the specific counties surrounding each of the airports mentioned earlier, and the area surrounding Washington DC itself. We further intend to use the cases and deaths to calculate the number of new cases and deaths occurring each day.

Data cleaning is performed using Python and Tableau Prep. Pandas library in python is used to drop the missing values and the redundant entries and remove the irrelevant columns from the datasets to decrease the processing time and enhance performance and efficiency. Moreover, we resolved the data inconsistencies, filtered the destination variable, and converted the UTC timestamp to date format using the Tableau Prep tool.

III. RELATED WORK

The rapid spread of coronavirus cases across the world motivates us to discover the number of flights effect on coronavirus deaths rate. As in almost every country, the first infection cases of coronavirus were brought by travelers. While travel restrictions have been applied in many countries, they had a modest effect on limiting the spread of coronavirus cases [35], [36]. These restrictions were effective in only delaying the transmission of coronavirus [37].

eone certainly has coronavirus. An additional lack of self-disclosure of coronavirus symptoms before and after boarding leads to an increase in the spread of the COVID-19 [29]. Previous studies have also reported that coronavirus can be transmitted before symptoms appear [30], [31], as people can be infected with coronavirus disease and show no symptoms, or have symptoms develop over a period of several days. Overall, Khanh et al. [29], Bae et al. [30], and Choi et al. [31] concluded that coronavirus could be transmitted on aircraft and consequently increase the infection risk.

A case study was applied in China to calculate the risk index of COVID-19 imported cases from inbound international flights [32]. Through this study, Zhang et al. [38] used global COVID-19 data and international flight data from UMETRIP, and they found that the risk index increases significantly when there are active flights associated with highly infected countries. A research on the effect of the United States local flights on COVID-19 cases indicates a high correlation, i.e., 0.8 between travelers and population and COVID-19 cases at the onset of the pandemic [39].

However, a study conducted and published in 2020 by Desmet and Wacziarg [40] used a cross-sectional regression model to analyze the relationship between COVID-19 cases and deaths and the distance to the closest airport with direct flights from the top five affected countries and found a negative correlation. Throughout this study, Desmet and Wacziarg [40] used the collected COVID-19 cases data from the New York Times and the international flights from the Bureau of Transportation Statistics. A retrospective case series was conducted by Yang et al. [41]. It was clinical data collected from ten patients with no symptom history before the flight. Yang et al. [41] found that coronavirus disease can be transmitted through airplanes and the way how transmitted was unknown. Similarly, a medical evaluation was conducted by Hoehl et al. [42] on twenty-four passengers from an international flight from Israel to Germany. Seven passengers were tested positive for coronavirus (SARS-CoV-2). Hoehl et al. [42] concluded that coronavirus transmission in an airplane depends on other elements such as passenger's movement, contact, etc. at the same time, wearing a mask during the flight could reduce the transition rate. On the other hand, a study by Schwartz et al. [43] stated that Coronavirus disease 2019 transmission was absent base on patients who traveled an internationalflight with one stop from China (Wuhan) to Canada (Toronto). Although several passengers developed some symptoms after the flight but tested negative from COVID-19.

Another study was conducted to investigate the association between air traffic volume and the spread of COVID-19. This study was conducted using the publicly available of domestic air traffic and passenger data from 2013 to 2018 through CAAC (Civil Aviation Administration of China). This data was used to predict the data for 2019. For the international air traffic and routes, data were derived from Chinese international air traffic from the Official Aviation Guide (OAG) and COVID-19 data from WHO [44]. They had found the continuous measurements using mean ± standard error, statistical significance using t-test, and correlation analysisusing linear regression. The analysis indicated a strong direct correlation between domestic covid cases and the number within China itself. The international air traffic analysis also showed a strong correlation with the cases as it depends on air traffic network

[44].

Furthermore, a mathematical modelling study was conducted to see whether the spread of COVID-19 was related to international cases imported. The study compared the ratio between the international imported cases and the internal cases in May 2020 and September 2020 in different countries [45]. Since each country has its own regulation implemented on the restriction, the results found at international cased imported made larger impact on May 2020 than September 2020. In fact, imported cases effect the internal spread in May 2020 was reduced compared to September 2020. However, the data collected for this study were from OpenSky data found the impact of international imported cases to internal spread is still little with 10% or less. [45]

## IV. METHODOLOGY

The first step in this project was cleaning the datasets.
Specifically, the Open Sky data, being crowd sourced, included errors such as duplicate entries or null origin and destination values among other things. This process enables us to continuously monitor the data quality and accuracy of our approach in model building. We further cut out the irrelevant data from the OpenSky dataset (that being the time period from 2019 until the start of the pandemic) as that information is irrelevant to our analysis. At this stage in the process, the coronavirus data was also filtered to only include the relevant region. Specifically, all cities and municipalities between the airports being studied and Washington, DC itself (including the counties containing each airport).

Our proposed method will involve marking all major airports in the United States and their immediately adjacent counties. We will then run a count on all flights arriving at each airport across the United States aggregated by week. Current CDC data suggests that the time from infection to onset of symptoms is 4 – 5 days [46] and research published in the Journal of Medical Virology indicates death from coronavirus has a median of 14.5 days after initial exposure [47]. As a result, based on this information, we intend to pair our flight data with coronavirus cases and deaths from one week and three weeks post the date of the flights. This should allow us to run several initial tests, including (assuming linear correlation) a Pearson Correlation Coefficient to determine if a correlation exists between number of flights and number of coronavirus deaths.

We can further take this data and produce scatterplots to visualize any correlation and determine other factors like spread and variance. Depending on the results of this analysis and visualization, we will then use machine learning in R and Python to experiment with appropriate regression methods and attempt to produce a model that can determine (with a degree of confidence) the likely number of additional coronavirus deaths that may result from the ongoing flights into a given airport. Control counties with 0 flights will be added to the model as well, chosen at random, which do not contain airports and are not adjacent to counties with airports.

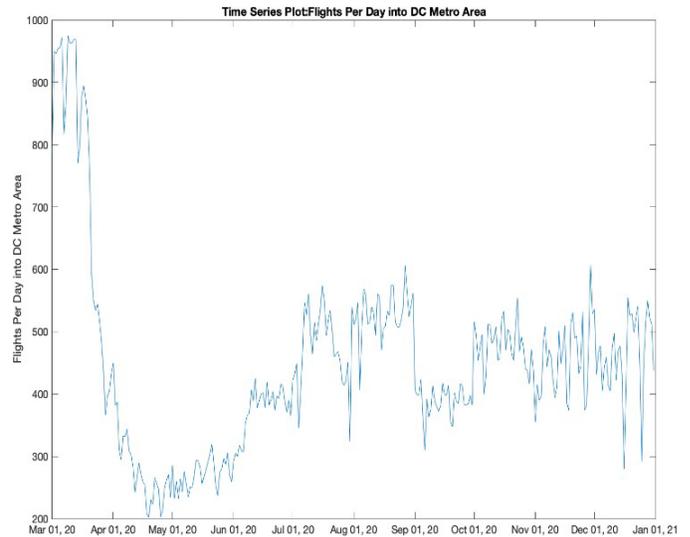

Fig. 1. Flights per day in DC Metro area

To allow for public review and availability of our data and ongoing progress, we will be posting the progress of our project online [48].

## V. INITIAL ANALYSIS AND VISUALIZATION

When the initial data preparation and cleaning were completed, we began using Pthon to visualize the data available. To begin, we took a look at the number of flights per day to determine if there are any clear trends, we should look at in the corresponding coronavirus data. As can be seen in Fig. 1 below, we were able to spot several regions of particularly low and high activity relative to the average of 447 flights per day.

Additionally, at this stage we began looking at the increases in cases and deaths per day for each region. Fig. 2 illustrates the cases and deaths reported in Washington DC itself for the time period of March 2020 through March of 2021 Noting that our flight data ends on January 1 st of 2021, there are still several clear peaks that can be looked at (at least in this data set) to determine if the number of flights into the DC region have an effect on the resulting cases/deaths.

To accomplish this initial look at how closely correlated flights are with deaths and cases per day, we ran some preliminary analysis and attempted to create scatterplots around our initial concept for the Washington DC area. That is to say that we compared flights with the number of cases reported a week later when we anticipated those would be detected or otherwise reported to the appropriate health agencies. When looking at this data we found unexpectedly that there was very limited negative correlation. Running Pearson's correlation coefficient against the two values yielded a result of -0.29 where the more flights reported the fewer cases were reported 7 days later. This can be roughly seen in Fig. 3 the scatterplot of these two values with a slight negative trend. Note however that the large accumulation of points near the right end of the graph at 1000 flights per day reflects early flight data towards the beginning of the pandemic and may need to be considered differently as lockdowns and containment procedures by airlines had not yet been implemented at the time.

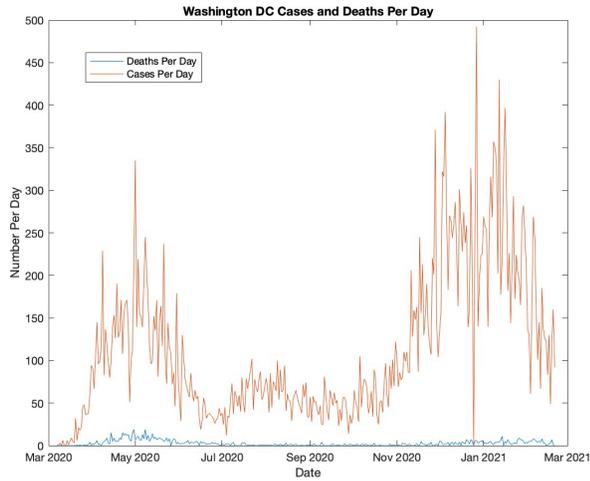

Fig. 2. Washington DC cases and death per day

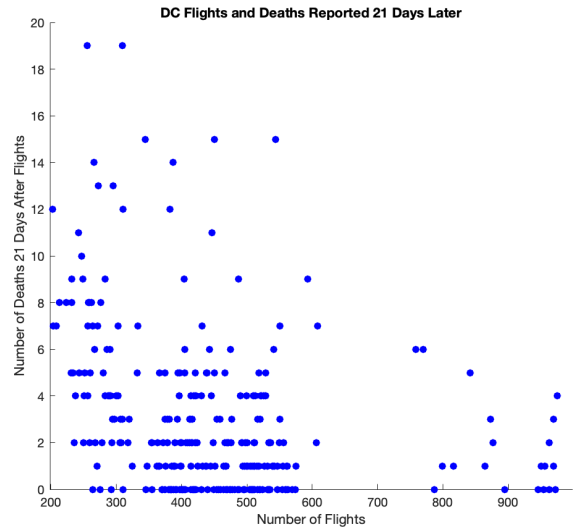

Fig. 4. Washington DC flights and deaths reported three weeks later

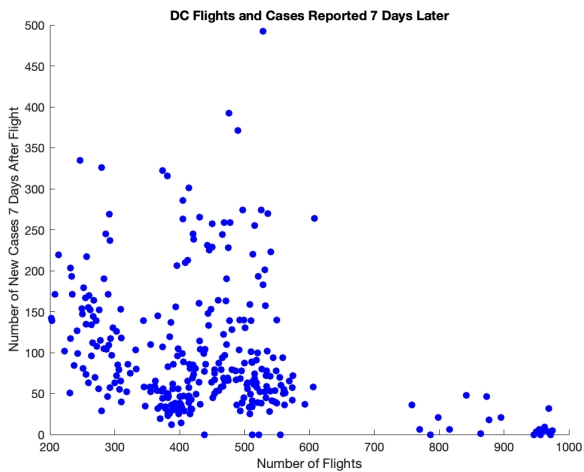

Fig. 3. Washington DC flights and cases reported a week later

Fig. 4 shows a slightly more pronounced negative correlation of -0.31 and a similarly more pronounced trend in the visible scatter plot of these values. Note that the same limitation of the cases applies, where the cluster of dots at the right side of the graph reflects data from early on in the pandemic before the coronavirus was as widespread.

As our team had initially hypothesized that there would be a clear positive correlation between these values, it's clear that these unanticipated results may require further review to ensure that our existing methodology is correct, particularly in light of the literature review that we previously conducted. It is however possible that our methodology is correct and that the negative correlation is due to other factors, such as local lockdowns, restrictions and policy which we also have yet to conclusively determine. Additionally, as noted before there is still data in our current analysis and visualizations reflecting the period early in the pandemic where the results of flight into the region may reflect differently on resulting cases than it would later when the virus is more common.

## VI. ANALYSIS

After reviewing the results of our initial analysis, we determined that starting our analysis on April 1st (after any government restrictions were put in place) would better reflect the impact of flights on case numbers. Additionally, as our initial method of looking at cases and deaths a number of days past the flight misses the large number of cases reported before and after the chosen day, we determined that a better method would be using an average of the number of new casesreported up to 14 days after the initial flight. Based on these updated methods, here were our analysis results:

### A. Loudoun County and Dulles International Airport

When looking at Dulles Airport and the surrounding Loudoun County, we determined that there was a fairly minimal but still present positive trend between the number of flights and new cases over 14 days, with a correlation coefficient of 0.16. In attempting to model this with linear regression however, the following result in Fig. 5 was produced.

The blue dots are a scatterplot of the actual data points, while yellow is the prediction based on the model. With 25% holdout validation, the root mean square error was 35.866.

Based on this, and the visible chart above, it seems apparent that while a positive correlation may exist, there is significant variance in the 14-days average of new cases that limits the predictive power of linear regression.

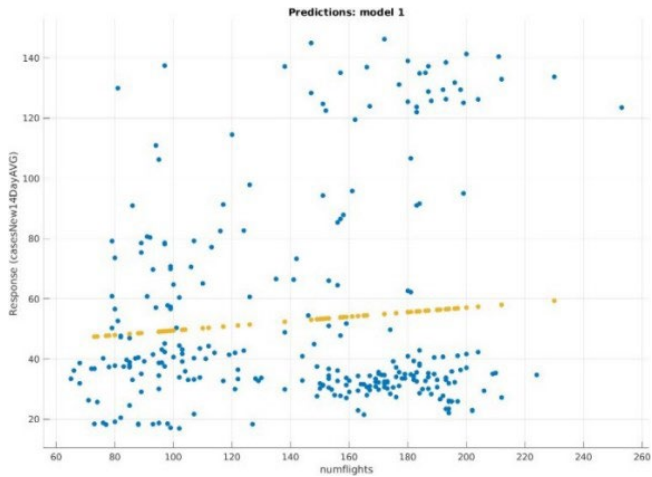

Fig. 5. Linear Regression Model- Flights into Dulles Airport and New Cases in Loudoun County over next 14 days.

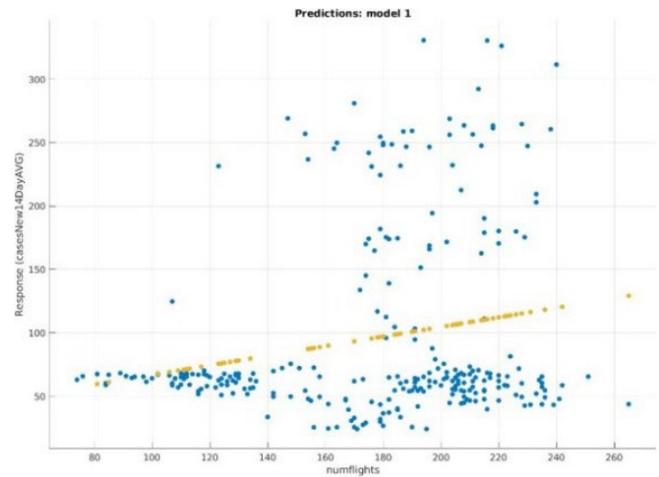

Fig. 7. Linear Regression Model- Flights into Baltimore International Airport and New Cases in Anne-Arundel County over next 14 days

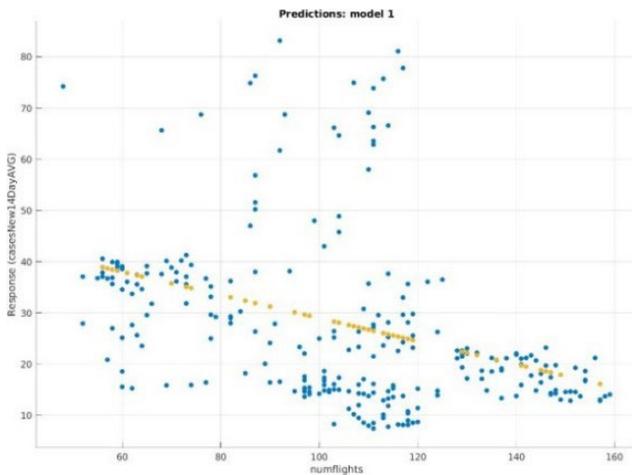

Fig. 6. Linear Regression Model- Flights into Reagan National Airport and New Cases in Arlington County over next 14 days

### B. Arlington and Reagan National Airport

A similar result was achieved when looking at Reagan National Airport and the surrounding Arlington County. Based on our data, we unexpectedly observed a negative trend between the number of flights and the number of new cases in Arlington, with a correlation coefficient of -0.376. Our linear correlation model is illustrated in Fig. 6. Based on the same 25% holdout validation, the root mean square error of this model was 15.084. While a much clearer negative trend exists in this chart, it should however be noted that there may be unaccounted confounding variables, such as the closure of Gate 35X at Reagan National Airport and subsequent opening of a new concourse over the next few months.

### C. Anne Arundel County and Baltimore Airport

Anne-Arundel County holds the last of the three major airports in the Washington DC Metro Area. Unlike Reagan National, this airport showed a slight positive correlation between the number of flights and the 14-day average of new cases, with a correlation coefficient of 0.2. Our resulting model can be seen in Fig. 7 below.

With 25% holdout validation, the root mean square error for this model was 41.43. As can be seen above, while there is a slight positive trend, most 14-day averages are clustered around 50 cases per day, regardless of the number of flights reported.

### D. Washington, DC itself

Among the considerations we had when looking at this data, we suspected that a large number of travelers into these airports may be flying into the region to work in Washington, DC as the largest city in the region. As a result, the final model we produced took all flights into all three airports and, like prior models, paired them to the 14-day average of new cases reported in Washington, DC. Our correlation coefficient showed a slight negative trend of -0.767, the specific model produced is illustrated below in Fig. 8.

As can be seen above, this was unfortunately lest predictive of our models. Based on 25% holdout validation, the RMSE of this model was a comparatively high 69.238.

## VII. CONCLUSIONS

What we determined based on our work in modeling the number of flights and resulting cases into the DC Metro Area is that the number of flights by itself seems to have a very minimal impact on the number of new cases reported over the next 14 days. We suspect this may be due to several reasons. It is possible that the number of flights (which notably non passenger flights carrying cargo) isn't a good metric for measuring how flights impact the number of reported cases in a region. It is also possible that airline policies have played a significant role in reducing in-flight transmission of coron avirus. We also surmise that a lack of testing requirements either before or after flights coupled with varying testing infrastructure in Maryland, Virginia, and Washington, DC may cause cases resulting from flight to be missed by our sources. Regardless, our conclusion at this time is that the number of flights by itself does not strongly correlate to the resulting number of new cases reported in the surrounding county over the next 14 days and should not be used as the sole predictor in future machine learning systems.

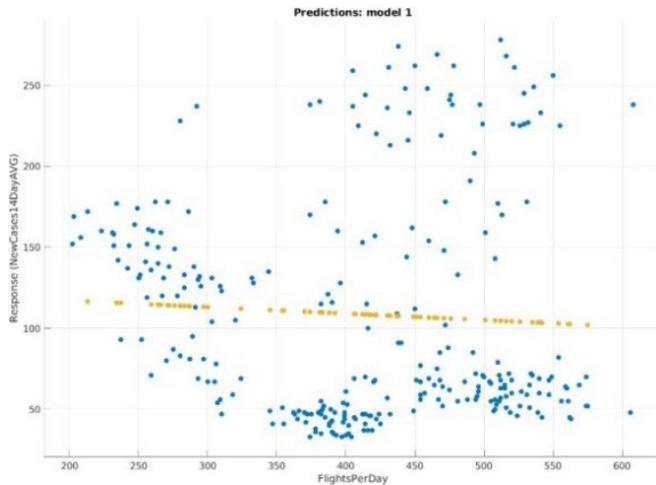

Fig. 8. Linear Regression Model- Flights into IAD, BWI, and DCA, and New Cases in the District of Columbia over the next 14 days

VIII. FUTURE WORK

Although the results showed little correlation because of travel by flight and COVID-19 cases in the DMV region, there are still significant improvements that can be made to our work. As COVID-19 is relatively still new, there are a lot of different experiments and tests that can be done. As a possible example, future algorithms may consider taking the origin airport into consideration. Another plausible research idea would be to see what kind of safety precautions were taken and using hypothesis testing to determine which specific measures reduce coronavirus transmission and to what degree. As a direct improvement to our method, rather than looking at the number of flights another approach for this project would havebeen to look at the total number of passengers each day, thus excluding the commercial cargo flights. This may allow for a clearer understanding of the correlation between air travel and COVID-19 cases. While it was deemed inappropriate for our usage (as we were anticipating and trying to determine linear correlation), yet another improvement would be attempting to model these data with an array of other unsupervised machine learning algorithms, selecting for the lowest RMSE to find a better model for the data (likely combined with other improvements mentioned above).

REFERENCES

[1] C. Assis, "Airline stocks slammed by coronavirus fears, but experts say reaction may be overdone." https://www.marketwatch.com/story/airline-stocks-slammed by-coronavirus-fears-but-experts-say-reaction-may-be-overdone-2020-03-06, March 7, 2020 (accessed April 25, 2021).
[2] C. Assis, "United airlines says coronavirus pandemic is worst crisis 'in the history of aviation." https://www.marketwatch.com/story/united-airlines-says-coronavirus-pandemic-is-worst-crisis-in-the-history-of-aviation-2020-04-30, May 1, 2020 (accessed April 26, 2021).
[3] J. Jolly, "Airlines may not recover from covid-19 crisis for five years, says airbus." https://www.theguardian.com/business/2020/apr/29/airlines-may-not-recover-from-covid-19-crisis-for-five-years-says-airbus, (accessed April 26, 2021).
[4] M. Heidari and J. H. Jones, "Using bert to extract topic-independent sentiment features for social media bot detection," in *2020 11th IEEE Annual Ubiquitous Computing, Electronics Mobile Communication Conference (UEMCON)*, pp. 0542–0547, 2020.
[5] M. Heidari, J. H. Jones, and O. Uzuner, "Deep contextualized word embedding for text-based online user profiling to detect social bots on twitter," in *2020 International Conference on Data Mining Workshops (ICDMW)*, pp. 480–487, 2020.
[6] M. Heidari and S. Rafatirad, "Semantic convolutional neural network model for safe business investment by using bert," in *2020 Seventh International Conference on Social Networks Analysis, Management and Security (SNAMS)*, pp. 1–6, 2020.
[7] M. Heidari, J. H. J. Jones, and O. Uzuner, "An empirical study of machine learning algorithms for social media bot detection," in *2021 IEEE International IOT, Electronics and Mechatronics Conference (IEMTRONICS)*, pp. 1–5, 2021.
[8] M. Heidari and S. Rafatirad, "Bidirectional transformer based on online text-based information to implement convolutional neural network model for secure business investment," in *2020 IEEE International Symposium on Technology and Society (ISTAS)*, pp. 322–329, 2020.
[9] S. Zad, M. Heidari, J. H. J. Jones, and O. Uzuner, "Emotion detection of textual data: An interdisciplinary survey," in *2021 IEEE World AI IoT Congress (AIIoT)*, pp. 0255–0261, 2021.
[10] M. Heidari and S. Rafatirad, "Using transfer learning approach to implement convolutional neural network model to recommend airline tickets by using online reviews," in *2020 15th International Workshop on Semantic and Social Media Adaptation and Personalization (SMA*, pp. 1–6, 2020.
[11] S. Zad, M. Heidari, J. H. Jones, and O. Uzuner, "A survey on concept-level sentiment analysis techniques of textual data," in *2021 IEEE World AI IoT Congress (AIIoT)*, pp. 0285–0291, 2021.
[12] M. Heidari, S. Zad, B. Berlin, and S. Rafatirad, "Ontology creation model based on attention mechanism for a specific business domain," in *2021 IEEE International IOT, Electronics and Mechatronics Conference (IEMTRONICS)*, pp. 1–5, 2021.
[13] M. Heidari, S. Zad, and S. Rafatirad, "Ensemble of supervised and unsupervised learning models to predict a profitable business decision," in *2021 IEEE International IOT, Electronics and Mechatronics Conference (IEMTRONICS)*, pp. 1–6, 2021.
[14] P. Hajibabaee, M. Malekzadeh, M. Ahmadi, M. Heidari, A. Esmaeilzadeh, R. Abdolazimi, and J. H. J. Jones, "Offensive language detection on social media based on text classification," in *2022 IEEE 12th Annual Computing and Communication Workshop and Conference (CCWC)*, pp. 0092–0098, 2022.
[15] S. Zad, M. Heidari, P. Hajibabaee, and M. Malekzadeh, "A survey of deep learning methods on semantic similarity and sentence modeling," in *2021 IEEE 12th Annual Information Technology, Electronics and Mobile Communication Conference (IEMCON)*, pp. 0466–0472, 2021.
[16] M. Heidari, J. H. J. Jones, and O. Uzuner, "Online user profiling to detect social bots on twitter," 2022.
[17] M. Heidari, S. Zad, P. Hajibabaee, M. Malekzadeh, S. HekmatiAthar, O. Uzuner, and J. H. Jones, "Bert model for fake news detection based on social bot activities in the covid-19 pandemic," in *2021 IEEE 12th Annual Ubiquitous Computing, Electronics Mobile Communication Conference (UEMCON)*, pp. 0103–0109, 2021.
[18] P. Hajibabaee, M. Malekzadeh, M. Heidari, S. Zad, O. Uzuner, and J. H. Jones, "An empirical study of the graphsage and word2vec algorithms for graph multiclass classification," in *2021 IEEE 12th Annual Information Technology, Electronics and Mobile Communication Conference (IEMCON)*, pp. 0515–0522, 2021.
[19] M. Malekzadeh, P. Hajibabaee, M. Heidari, S. Zad, O. Uzuner, and J. H. Jones, "Review of graph neural network in text classification,"

in *2021 IEEE 12th Annual Ubiquitous Computing, Electronics Mobile Communication Conference (UEMCON)*, pp. 0084–0091, 2021.
[20] R. Abdolazimi, M. Heidari, A. Esmaeilzadeh, and H. Naderi, "Mapreduce preprocess of big graphs for rapid connected components detection," in *2022 IEEE 12th Annual Computing and Communication Workshop and Conference (CCWC)*, pp. 0112–0118, 2022.
[21] A. Esmaeilzadeh, M. Heidari, R. Abdolazimi, P. Hajibabaee, and M. Malekzadeh, "Efficient large scale nlp feature engineering with apache spark," in *2022 IEEE 12th Annual Computing and Communication Workshop and Conference (CCWC)*, pp. 0274–0280, 2022.
[22] S. Rafatirad and M. Heidari, "An exhaustive analysis of lazy vs. eager learning methods for real-estate property investment," 2019.
[23] M. Malekzadeh, P. Hajibabaee, M. Heidari, and B. Berlin, "Review of deep learning methods for automated sleep staging," in *2022 IEEE 12th Annual Computing and Communication Workshop and Conference (CCWC)*, pp. 0080–0086, 2022.
[24] M. Heidari and J. H. J. Jones, "Bert model for social media bot detection," 2022.
[25] M. Heidari, "Nlp approach for social media bot detection(fake identity detection) to increase security and trust in online platforms," 2022.
[26] J. Read, "How clean is the air on planes?." https://www.nationalgeographic.com/travel/article/how-clean-is-the-air-on-your-airplane-coronavirus-cvd, August 28, 2020 (accessed April 27, 2021).
[27] Centers for Disease Control and Prevention (CDC), "How covid-19 spreads." https://www.cdc.gov/coronavirus/2019-ncov/prevent-getting-sick/how-covid-spreads.html, July 14, 2021.
[28] R. J. Milne, C. Delcea, and L.-A. Cotfas, "Airplane boarding methods that reduce risk from covid-19," *Safety Science*, vol. 134, p. 105061, 2021.
[29] N. C. Khanh, P. Q. Thai, H.-L. Quach, N.-A. H. Thi, P. C. Dinh, T. N. Duong, L. T. Q. Mai, N. D. Nghia, T. A. Tu, L. N. Quang, T. D. Quang, T.-T. Nguyen, F. Vogt, and D. D. Anh, "Transmission of SARS-CoV 2 during long-haul flight," *Emerging Infectious Diseases*, vol. 26, pp. 2617–2624, Nov. 2020.
[30] S. H. Bae, H. Shin, H.-Y. Koo, S. W. Lee, J. M. Yang, and D. K. Yon, "Asymptomatic transmission of SARS-CoV-2 on evacuation flight," *Emerging Infectious Diseases*, vol. 26, pp. 2705–2708, Nov. 2020.
[31] E. M. Choi, D. K. Chu, P. K. Cheng, D. N. Tsang, M. Peiris, D. G. Bausch, L. L. Poon, and D. Watson-Jones, "In-flight transmission of SARS-CoV-2," *Emerging Infectious Diseases*, vol. 26, pp. 2713–2716, Nov. 2020.
[32] Centers for Disease Control and Prevention (CDC), "First travel-related case of 2019 novel coronavirus detected in united states." https://www.cdc.gov/media/releases/2020/p0121-novel-coronavirus-travel-case.html, (accessed April 27, 2021).
[33] X. Olive, M. Strohmeier, and J. Lübbe, "Crowdsourced air traffic data from the opensky network 2020," 2021.
[34] The New York Times, "Nytimes/covid-19-data." https://github.com/nytimes/covid-19-data/blob/master/us-states.csv, 2021, March 13.
[35] M. Bielecki, D. Patel, J. Hinkelbein, M. Komorowski, J. Kester, S. Ebrahim, A. J. Rodriguez-Morales, Z. A. Memish, and P. Schlagenhauf, "Air travel and COVID-19 prevention in the pandemic and peri-pandemic period: A narrative review," *Travel Medicine and Infectious Disease*, vol. 39, p. 101915, Jan. 2021.
[36] M. Chinazzi, J. T. Davis, M. Ajelli, C. Gioannini, M. Litvinova, S. Merler, A. P. y Piontti, K. Mu, L. Rossi, K. Sun, C. Viboud, X. Xiong, H. Yu, M. E. Halloran, I. M. Longini, and A. Vespignani, "The effect of travel restrictions on the spread of the 2019 novel coronavirus (COVID-19) outbreak," *Science*, vol. 368, pp. 395–400, Mar. 2020.
[37] A. Adekunle, M. Meehan, D. Rojas-Alvarez, J. Trauer, and E. McBryde, "Delaying the COVID-19 epidemic in australia: evaluating the effectiveness of international travel bans," *Australian and New Zealand Journal of Public Health*, vol. 44, pp. 257–259, July 2020.
[38] L. Zhang, H. Yang, K. Wang, Y. Zhan, and L. Bian, "Measuring imported case risk of COVID-19 from inbound international flights — a case study on china," *Journal of Air Transport Management*, vol. 89, p. 101918, Oct. 2020.
[39] J. A. Ruiz-Gayosso, M. del Castillo-Escribano, E. Hernández-Ramírez, M. del Castillo-Mussot, A. Pérez-Riascos, and J. Hernández-Casildo, "Correlating USA COVID-19 cases at epidemic onset days to domestic flights passenger inflows by state," *International Journal of Modern Physics C*, vol. 32, p. 2150014, Nov. 2020.
[40] K. Desmet and R. Wacziarg, "Understanding spatial variation in COVID-19 across the united states," tech. rep., National Bureau of Economic Research, June 2020.
[41] N. Yang, Y. Shen, C. Shi, A. H. Y. Ma, X. Zhang, X. Jian, L. Wang, J. Shi, C. Wu, G. Li, Y. Fu, K. Wang, M. Lu, and G. Qian, "In-flight transmission cluster of COVID-19: A retrospective case series," Mar. 2020.
[42] S. Hoehl, O. Karaca, N. Kohmer, S. Westhaus, J. Graf, U. Goetsch, and S. Ciesek, "Assessment of SARS-CoV-2 transmission on an international flight and among a tourist group," *JAMA Network Open*, vol. 3, p. e2018044, Aug. 2020.
[43] K. L. Schwartz, M. Murti, M. Finkelstein, J. A. Leis, A. Fitzgerald-Husek, L. Bourns, H. Meghani, A. Saunders, V. Allen, and B. Yaffe, "Lack of COVID-19 transmission on an international flight," *Canadian Medical Association Journal*, vol. 192, pp. E410–E410, Apr. 2020.
[44] H. Lau, V. Khosrawipour, P. Kocbach, A. Mikolajczyk, H. Ichii, M. Zacharski, J. Bania, and T. Khosrawipour, "The association between international and domestic air traffic and the coronavirus (COVID-19) outbreak," *Journal of Microbiology, Immunology and Infection*, vol. 53, pp. 467–472, June 2020.
[45] T. W. Russell, J. T. Wu, S. Clifford, W. J. Edmunds, A. J. Kucharski, and M. Jit, "Effect of internationally imported cases on internal spread of COVID-19: a mathematical modelling study," *The Lancet Public Health*, vol. 6, pp. e12–e20, Jan. 2021.
[46] CDC, "Interim clinical guidance for management of patients with confirmed coronavirus disease (covid-19)." https://www.cdc.gov/coronavirus/2019-ncov/hcp/clinical-guidance-management-patients.html, 16 Feb, 2021.
[47] W. Wang, J. Tang, and F. Wei, "Updated understanding of the outbreak of 2019 novel coronavirus (2019-nCoV) in wuhan, china," *Journal of Medical Virology*, vol. 92, pp. 441–447, Feb. 2020.
[48] M. Thompson, G. Alshabana, T. Tran, and A. Chitimalla, "Predict covid-19 cases using opensky data." http://mason.gmu.edu/ttran81/, 2021.